\documentclass{article}

\PassOptionsToPackage{numbers, compress}{natbib}
\usepackage[preprint]{neurips_2026}

\usepackage[utf8]{inputenc} 
\usepackage[T1]{fontenc}    
\usepackage{wrapfig}
\usepackage{hyperref}       
\usepackage{url}            
\usepackage{booktabs}       
\usepackage{amsfonts}       
\usepackage{nicefrac}       
\usepackage{microtype}      
\usepackage{xcolor}         

\title{Do vision models know that three is three?}

\author{
  \makebox[\dimexpr\textwidth-2\tabcolsep\relax][c]{
    \begin{tabular}{@{}c@{}}
      Ankit Sonthalia$^{1}$ \quad Arnas Uselis$^{1}$ \quad Seong Joon Oh$^{2}$
  \end{tabular}}
}

\usepackage{aliascnt}
\usepackage{array}
\usepackage{amsmath}
\usepackage{amsthm}
\usepackage[capitalize,noabbrev]{cleveref}
\usepackage{graphicx}
\usepackage{subcaption}
\usepackage{longtable}

\theoremstyle{plain}

\theoremstyle{definition}
\newtheorem{definition}{Definition}[section]

\theoremstyle{remark}

\newcommand{\quotes}[1]{``#1''}

\begin{document}

\newcommand{\NUMMODELS}{46}
\newcommand{\NUMCONCEPTFAMILIES}{10}

\newcommand{\conceptspace}{\mathcal{Z}}
\newcommand{\partition}{\mathcal{P}}

\newcommand{\RANDOMCURVECOLOR}{888888}
\newcommand{\LINEARITYCURVECOLOR}{2A7D72}

\newcommand{\LocalDCViTSupMNISTCount}{0.45}
\newcommand{\LocalDCViTSupMNISTCountPVal}{4.1 \times 10^{-73}}
\newcommand{\LocalDCViTSupRoomCleanliness}{0.52}
\newcommand{\LocalDCViTSupRoomCleanlinessPVal}{1.3 \times 10^{-26}}
\newcommand{\LocalLinearityPct}{99.1}
\newcommand{\LocalLinearityTests}{2{,}530}
\newcommand{\LocalLinearityFloorPctDryWet}{97.8}

\newcommand{\GlobalLinearityPctObjects}{99.3}
\newcommand{\GlobalLinearityPctAge}{94.6}

\newcommand{\GlobalLinearityPct}{52.2}
\newcommand{\GlobalLinearityTests}{13{,}984}

\newcommand{\FragViTSupCounting}{0.67}
\newcommand{\FragInsigViTSupCounting}{8}
\newcommand{\FragViTSupCleanliness}{0.92}
\newcommand{\FragInsigViTSupCleanliness}{11}
\newcommand{\FragCLIPCleanliness}{0.08}

\newcommand{\ProbeTauCount}{0.77}
\newcommand{\ProbeTauCleanliness}{0.66}
\newcommand{\ProbeTauSunny}{0.53}

\maketitle

{\renewcommand{\thefootnote}{}\footnotetext{$^{1}$University of T\"ubingen. $^{2}$KAIST AI. Correspondence to Ankit Sonthalia \textless{}ankit.sonthalia@uni-tuebingen.de\textgreater{}.}}

\begin{abstract}

Do vision models encode the transition $2{\to}3$ the same way whether it appears in handwritten digits or in hand gestures, and do they \quotes{understand} that clean${\to}$dirty is the same transformation regardless of whether it is applied to a kitchen or a carpet? The linear representation hypothesis (LRH) answers this for language models: concept changes correspond to directions in (language) embedding space. In vision, the same concept can manifest in visually unrelated \emph{regimes} that rarely co-occur: images of digits and hands do not share much in common in pixel space. To understand whether vision models still encode the underlying abstract concepts, we study $\NUMMODELS$ vision encoders across architectures, training objectives, and scales, on $\NUMCONCEPTFAMILIES$ concept families including numerosity, age, object identity, and MIT-States attributes. Within each regime, we find statistically significant concept directions near-universally across our tested models. However, two distinct regimes do not always share the same concept direction. Several models learn multiple linear structures for the same concept transition, split across regimes. We call this phenomenon \emph{fragmented linearity}: strong local linearity coexisting with near-absent global linearity. Fragmentation is, however, not fixed: models with stronger local linearity progressively escape it (local-global linearity correlated with $\tau \in [0.38, 0.89]$ across all concepts), and where global linearity does emerge, concept directions can transfer as linear probes across regimes.

\end{abstract}

\section{Introduction}
\label{sec:intro}

\begin{figure}[t]
    \centering
    \captionsetup[subfigure]{skip=3pt}
    \begin{subfigure}[t]{0.48\linewidth}
        \centering
        \includegraphics[width=\linewidth]{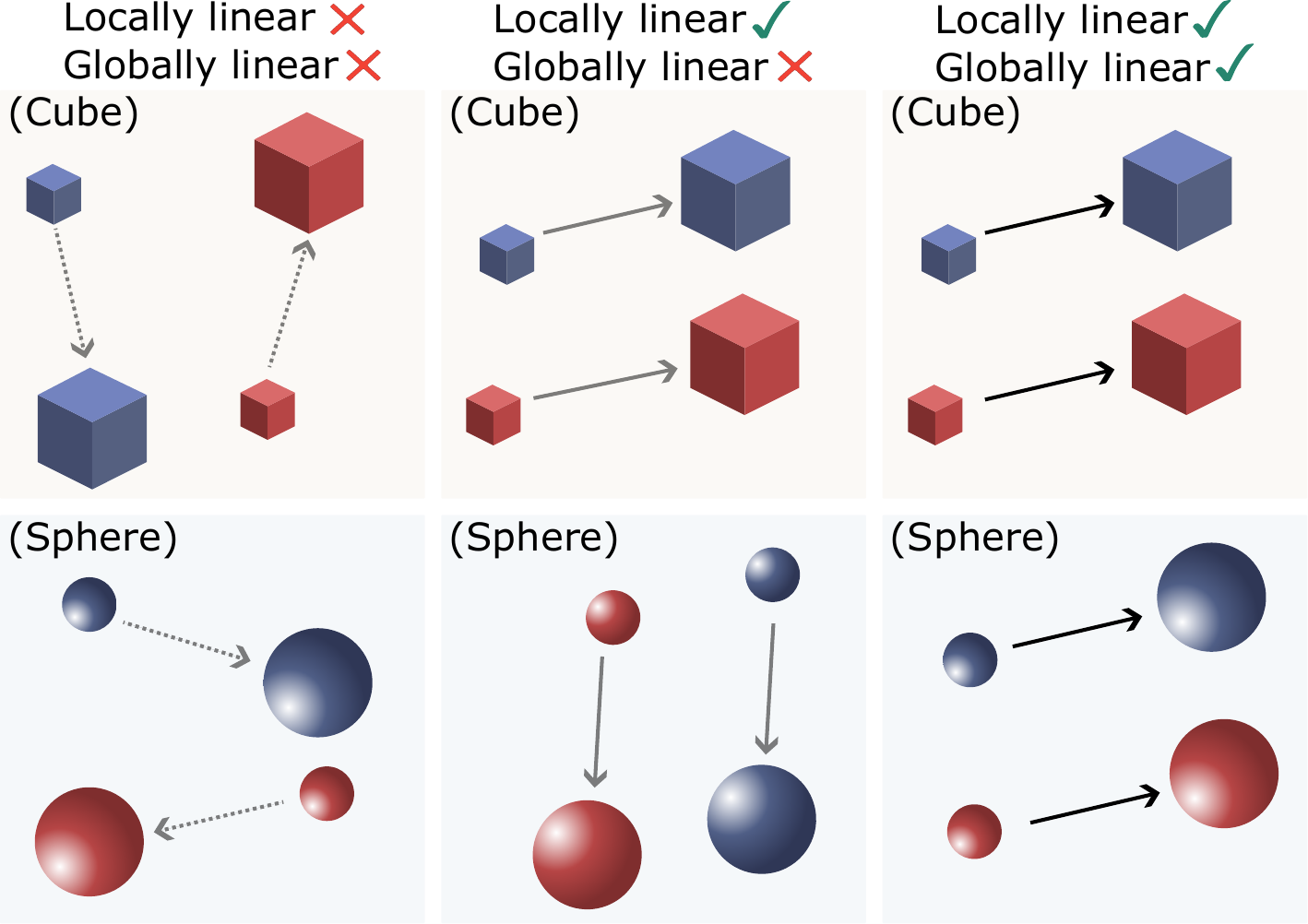}
        \subcaption{Schematic: Is \quotes{small $\rightarrow$ big} encoded as a single direction across visual regimes (e.g., cubes vs.\ spheres)?
        \textbf{Left}: no consistent direction within any regime.
        \textbf{Middle}: each regime has a consistent direction, but directions differ across regimes (\textit{fragmented linearity}).
        \textbf{Right}: one direction holds globally.
        }
    \end{subfigure}\hspace{0.015\linewidth}%
    \begin{subfigure}[t]{0.48\linewidth}
        \centering
        \includegraphics[width=\linewidth]{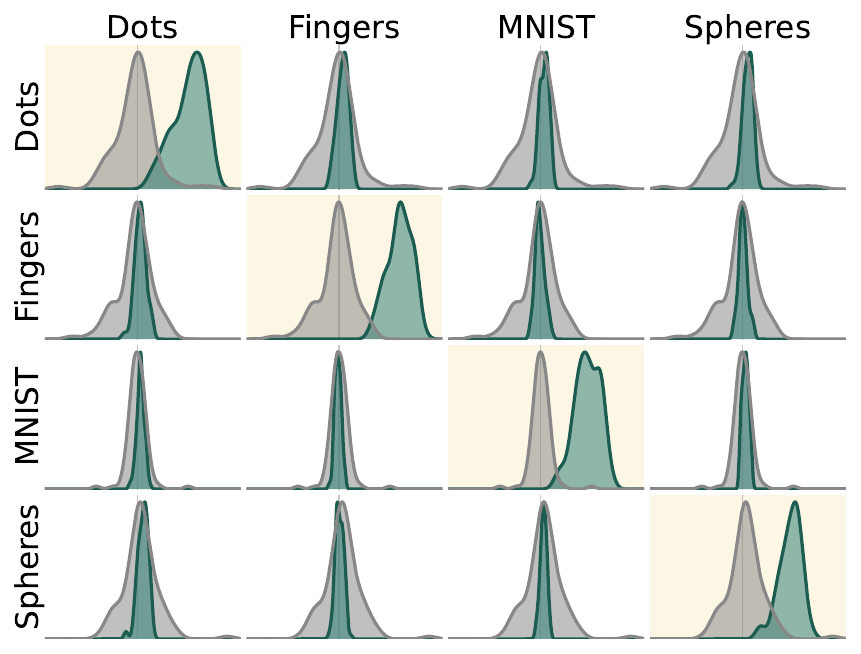}
        \subcaption{[ViT-L/16-Sup] KDEs of alignment between discovered concept directions in regime $i$ (rows) and \textcolor[HTML]{\LINEARITYCURVECOLOR}{\textbf{actual counterfactual differences}} in regime $j$ (columns) vs.\ \textcolor[HTML]{\RANDOMCURVECOLOR}{\textbf{random vectors}} ($2{\to}3$ transition): within-regime (on-diagonal) cosines exceed the baseline; cross-regime (off-diagonal) they typically do not.
        }
    \end{subfigure}
    \caption{\textbf{Vision embeddings exhibit fragmented linearity.} (a) Three possibilities for linearity: none, fragmented, or global. (b) Empirically, we often observe fragmented linearity.}
    \label{fig:linearity}
    \vspace{-1.3em}
\end{figure}

The linear representation hypothesis (LRH) \cite{mikolov2013linguisticregularitiescontinuous, park2024linear} claims that the embedding spaces of large language models (LLMs) contain linear directions corresponding to high-level concept changes, enabling interpretability and algebraic interventions in representation space. 
Yet as vision encoders increasingly serve as the perceptual backbone of multimodal systems \cite{liu_visual_2023, karamcheti2024prismatic}, understanding whether the same linear structure holds in vision is essential to completing the picture.

Vision is fundamentally different from language in that the same concept often manifests in visually drastically different ways. 
For instance, the concept of \quotes{3} can be represented using handwritten digits \cite{li_deng_mnist_2012}, the number of objects in an image \cite{johnson2016clevr} or hand gestures \cite{koryakin_fingers}. While these different groups (\emph{regimes}) of images do not share much in common in pixel space, a human would understand that the same underlying concept is being manipulated; do vision embedding spaces encode the same abstractions? A linear representation in embedding space would hint that they do.

Linear structure (\emph{linearity}) in the embedding space is indeed often an assumption (implicit or otherwise) in prior works that attempt to interpret vision embeddings \cite{kim_interpretability_2018, bhalla_interpreting_2024, thasarathan_universal_2025}. 
Prior work has also studied linearity of vision embeddings after projection onto a tangent space \cite{berasi_not_2025}, with respect to ordinal attributes \cite{sonthalia_rankability_2025}, and its emergence as a consequence of data diversity \cite{uselis_does_2025}.
While these findings hint toward the presence of linear structure in vision embeddings, a more rigorous study exploring whether it survives across drastically unrelated visual regimes remains missing. 

To address this gap, we study \NUMMODELS\ vision encoders varying in architecture, training objective, model size and pretraining data, using the concepts of numerosity, age, and object identity, as well as $7$ attributes curated from MIT-States \cite{isola2015discoveringstatestransformations}. We show that within a given regime, concepts are almost universally encoded linearly to an extent that is statistically distinguishable from a random baseline. However, this applies only partially to cross-regime transfer (\cref{fig:linearity}). This finding motivates our definitions of \textit{local} (within-regime), \textit{global} (cross-regime) and \textit{fragmented} (only within-regime without cross-regime) linearity. 

While local linearity is present even in supervised models with complete fragmentation across regimes (supervised models), models with stronger local linearity (CLIP-like models) also tend to exhibit stronger global linearity. This suggests a gradual convergence of regime-specific representations into abstract regime-invariant ones. 

Our contributions are:

\begin{enumerate}
    \item A systematic investigation of the LRH for vision models.
    \item Statistically rigorous empirical evidence that abstract concepts exhibit \textit{fragmented linearity}. 
    \item Empirical evidence that modern vision encoders exhibit increasingly globally linear concept directions.
\end{enumerate}

\section{Preliminaries and related work}
\label{sec:prelim_related}

We review three different notions of linear structure that appear in prior work and situate our study within the existing landscape. Of these, the LRH is the main notion studied in our work. 

\textbf{Notation.}
We assume a concept space $\mathcal{Z} = Z_1 \times Z_2 \times \cdots \times Z_k$ 
where the $i$th concept can take one of $\mid Z_i \mid$ values; 
samples $x \in \mathcal{X}$ are generated as $x = g(z), z \in \mathcal{Z}$ 
where $\mathcal{X}$ is the input space (images or text prompts).  
An (image or text) encoder is a function $f: \mathcal{X} \rightarrow \mathbb{R}^D$. 
Where convenient, we will use the notation $\phi(z) := f(g(z))$ for some $z \in \mathcal{Z}$. 

\subsection{Different notions of linearity}
\label{sec:different_notions_linearity}

\paragraph{(1) Linear separability.}
Linear probes \cite{alain_understanding_2018} reveal whether different classes of a $\mid Z_i \mid$-valued concept can be separated linearly using a classifier 
$h : \mathbb{R}^D \rightarrow [0,1]^{\mid Z_i \mid}$.
This notion of linearity has been largely shown to hold for many modern models \cite{chen2020simpleframeworkcontrastive, he2020momentumcontrastunsupervised, radford2021learning, oquab2024dinov2, uselis_intermediate_2025, vompa2025linearseparabilityceiling}; however, separability does not guarantee that concepts admit consistent linear directions conducive to manipulation. 

\paragraph{(2) Linear representation hypothesis (LRH).}

According to the LRH \cite{park2024linear}, concepts are represented as \textit{directions} in embedding space. Formally, for two values $a, b \in Z_i$ of the $i$-th concept, define a set of counterfactual pairs $\mathcal{C}_{a \rightarrow b}^{\mathcal{Z}, i}$ where samples differ only in the value of this concept:

\begin{equation}
\label{eq:counterfactual}
\mathcal{C}_{a \rightarrow b}^{\mathcal{Z}, i} = \{ (z, z') 
\in \mathcal{Z} \times \mathcal{Z} \mid z_i = a,\; z'_i=b,\; a\neq b,\; z_j=z'_j \ \forall j\neq i\}.
\end{equation}

\begin{definition}[Directional linearity (LRH) \cite{park2024linear}]
\label{def:lrh}

We define concept $i$ as being represented \textbf{directionally linearly} in $f$ if for every $a, b \in Z_i$, there exists a fixed $v_{a \rightarrow b}^{\mathcal{Z}, i} \in \mathbb{R}^D$ such that for every counterfactual pair $(z, z') \in \mathcal{C}_{a \rightarrow b}^{\mathcal{Z}, i}$, there exists $\alpha \in \mathbb{R}$ such that 
\[
\exists \alpha \; \text{s.t.} \;\phi(z') - \phi(z) = \alpha v_{a \rightarrow b}^{\mathcal{Z}, i} 
\]

We refer to $v_{a \rightarrow b}^{\mathcal{Z}, i}$ as the \textbf{concept direction} from $a$ to $b$.

\end{definition}

\paragraph{(3) Linear additive decomposability.}
\cite{trager_linear_2024} define linearity as the property where primitive concepts can be represented as vectors that yield composite concepts when added together. We refer to this notion of linearity as \textit{additive} linearity.

\begin{definition}[Additive linearity \cite{trager_linear_2024}]
\label{def:additive_linearity}
 
We define $f$ to be \textbf{additively linear} if the following holds for every $i$: for every $a, b \in Z_i$, there exists a fixed $v_{a \rightarrow b}^{\mathcal{Z}, i} \in \mathbb{R}^D$ such that for every counterfactual pair $(z, z') \in \mathcal{C}_{a \rightarrow b}^{\mathcal{Z}, i}$,
\[
\phi(z') - \phi(z) = v_{a \rightarrow b}^{\mathcal{Z}, i}.
\]

\end{definition}

\cite{berasi_not_2025} show additive linearity holds approximately after tangent space projection, but it remains unclear (i) whether the weaker LRH still holds in Euclidean space and (ii) whether any linear structure transfers across visually disjoint regimes. Our study aims to address both.

\subsection{Related work}

\paragraph{Interpretability and the linearity assumption.}
Early interpretability research pioneered by \cite{kim_interpretability_2018} assumes local linear separability of samples belonging to a concept from those not belonging to it. The assumption of concepts lying in some low-dimensional subspace is implicit \cite{yeh_completeness-aware_2020} but not otherwise verified. 
More recent works \cite{gorton_missing_2024, bhalla_interpreting_2024, thasarathan_universal_2025} train SAEs to sparsely decompose embeddings. Nevertheless, it is unclear whether SAEs recover the \quotes{true} concept directions \cite{fel_archetypal_2025}. While the goal in these works is interpretability of vision models, we focus on characterizing the geometry that enables interpretability. 

\textbf{Linear structure in vision models.}
\cite{wolff_independent_2023} discover task-aligned subspaces in CLIP's image encoder.
Extending the findings from \cite{trager_linear_2024} to the vision domain, \cite{berasi_not_2025} show that CLIP image embeddings admit approximately linear structure after projection onto the tangent space. 
\cite{sonthalia_rankability_2025} show that many ordinal attributes can be linearly decoded from popular vision encoders, while \cite{uselis_does_2025} show compositional linearity emerges as a consequence of training data diversity. 

The above works assess linearity as a single global property of the representation. 
In contrast, we consider visually different regimes separately and hypothesize that (1) individual regimes contain strong concept representations, and (2) linearity (\cref{def:lrh}) often holds only weakly because regime-specific concept representations fail to transfer strongly. This is also akin to learning different mechanisms for solving the same task in different contexts \cite{kumar_questioning_2025}, although mechanistic interpretability is outside our scope. 

\textbf{VLM embedding spaces.}
A broad body of prior work studies different aspects of VLM embedding spaces in general. Of particular note have been the modality gap
\cite{schrodi_two_2025, liang_mind_2022, levi2024doubleellipsoid}, binding capabilities \cite{lewis_does_2024, koishigarina2025clip, al-tahan_scaling_nodate} and negation \cite{ranjbar_spacevlm_2025}. \cite{kang_is_2025} propose geometric constraints for an \quotes{ideal} CLIP. In contrast to these works, our work explores the local geometry of general vision models, including CLIP.

\textbf{Compositional generalization and compositional zero-shot learning.} 
VLMs have been frequently shown to be suitable for compositional zero-shot learning \cite{nayak_learning_2022, perera_prompt_2023, qu_learning_2025}, compositional retrieval \cite{baldrati_zero-shot_2023, karthik_vision-by-language_2024, wang_generative_nodate}, and concept editing \cite{ceylan_pix2video_2023}. Other work has revealed shortcomings \cite{tong_eyes_2024}. We take a complementary approach and analyze the \textit{geometry} of existing vision encoders for local linearity; our findings may contribute towards explaining the compositional failures of VLMs.

\textbf{Representation convergence.} The Platonic Representation Hypothesis \cite{huh_platonic_2024} argues that learned representations are converging toward a shared, canonical model of the world. In contrast, we study a different form of convergence: whether a single model’s embedding space aligns \emph{the same concept across different regimes}. Our perspective is closer to work questioning whether neural networks reuse identical computational structures across settings \cite{lepori_break_2023, kumar_questioning_2025, mccracken_uncovering_2025}. In contrast, however, we focus on the final-layer embeddings which are directly exposed to downstream tasks like zero-shot retrieval and classification.

\vspace{-0.5em}
\section{Fragmented Linearity}
This section defines local linearity and fragmentation and lays out the measurements used throughout the paper.

\subsection{Regimes and approximate counterfactuals}

Our linearity metric follows the LRH definition in \cref{def:lrh}.

\subsubsection{Regimes and local linearity}
\label{sec:regimes}
Consider the texts \quotes{3 fingers} and \quotes{the digit 3}: in language, the shared concept \quotes{3} is explicit. 
No comparable explicit signal links an image of the digit 3 to an image of three raised fingers. 
Similarly, \quotes{old coin} and \quotes{old building} make the connection explicit but images of old coins and old buildings do not.
This motivates studying concept structure in vision at the level of data subsets (digits, fingers, spheres, coins, buildings, and so on), which we call \emph{regimes}. 
Formally, we use a partition $\partition=\{S_r\}_{r=1}^m$ with $S_r \subseteq \conceptspace$ and $\bigcup_{r=1}^m S_r=\conceptspace$, and call each $S_r$ a regime.
This allows us to study linearity within and across regimes.

\begin{definition}[Local linearity]
\label{def:local_linearity}
Given a concept space $\conceptspace$ and a partition $\partition=\{S_i\}_{i=1}^m$ with $\bigcup_i S_i=\conceptspace$, we define $f$ to be \textbf{locally linear with respect to $\partition$} (under linearity notion $L$) if $L$ holds inside each $S_i\in\partition$.
\end{definition}

Note that if $L$ holds for all $\partition\in \Pi(\conceptspace)$, then $f$ is globally linear since $\{\conceptspace\} \in \Pi(\conceptspace)$.

\subsubsection{Counterfactual construction}
\label{sec:counterfactuals}

In \cref{eq:counterfactual}, counterfactual pairs differ only in one concept. 
Because our analysis is at the regime level, we construct counterfactuals within a fixed regime $S_r$. 

Images are sparse in concept space even within regimes. A handwritten \quotes{2} and a handwritten \quotes{3} differ in other concepts like style and stroke width; one can rarely match all such concepts across two images (text avoids this problem by abstracting these concepts away). 
We therefore construct \emph{approximate} counterfactual pairs for transition $a \rightarrow b$ of concept $i$ inside each regime.

Fix a regime $S_r$. We form two finite embedding sequences within this regime: one each for samples with the $i$-th concept set to $a$ and $b$ respectively. Let $E=(e_k)_{k=1}^n$ with $e_k\in\{\phi(z)\mid z\in S_r,\; z_i=a\}$, and let $E'=(e'_\ell)_{\ell=1}^n$ with $e'_\ell\in\{\phi(z')\mid z'\in S_r,\; z'_i=b\}$. 
We compute a bijection $\pi^*$ by solving the linear sum assignment problem
\begin{equation}
    \pi^* = \arg\min_{\pi \in S_n} \sum_{k=1}^{n} \|e_k-e'_{\pi(k)}\|_2.
\end{equation}

The matched embedding pairs $\{(e_k, e'_{\pi^*(k)})\}_{k=1}^n$ define our approximation to $\mathcal{C}_{a \rightarrow b}^{S_r, i}$. Under the assumption that embedding distance reflects similarity in non-target concepts, this construction yields the closest available counterfactual pairs within regime $S_r$.

\subsection{Metrics: direction consistency and fragmentation}

We introduce our linearity metric (direction consistency) and distinguish local linearity from global linearity. This then motivates our fragmentation metric.

\subsubsection{Direction consistency}

Direction consistency (DC) measures whether counterfactual differences for a concept align with a single common direction (\cref{def:lrh}, \cite{park2024linear}).

\begin{definition}[Within-regime direction consistency]
\label{def:direction_consistency_local}
This measures whether a given regime contains a coherent concept direction.
Fix concept $i$ and values $a,b\in Z_i$. Let $\partition=\{S_r\}_{r=1}^m$ and fix $S\in\partition$. Let $\mathcal{C}_{a\rightarrow b}^{S,i}=\{(z_k,z_k')\}_{k=1}^{n_S}$ be matched counterfactual pairs and $\Delta_k^{S,i}:=\phi(z_k')-\phi(z_k)$. For each held-out index $k$, define
\[
\hat v_{a\rightarrow b}^{S,-k,i}:=\frac{1}{n_S-1}\sum_{j\neq k}\Delta_j^{S,i},
\]
The within-regime DC score averages leave-one-out cosine alignment between each held-out displacement $\Delta_k^{S,i}$ and its leave-one-out estimated concept direction $\hat v_{a\rightarrow b}^{S,-k,i}$.
\begin{equation}
\label{eq:dc_local}
\mathrm{DC}_{a\rightarrow b}^{i}(S):=
\frac{1}{n_S}\sum_{k=1}^{n_S}
\cos\!\left(\Delta_k^{S,i},\hat v_{a\rightarrow b}^{S,-k,i}\right).
\end{equation}
We call $\mathrm{DC}_{a\rightarrow b}^{i}(S)$ \textbf{local DC}.
High local DC for all $S\in\partition$ implies high local linearity with respect to $\partition$.
\end{definition}

Given a partition $\partition=\{S_r\}_{r=1}^m$, we summarize local linearity by averaging local DC across regimes:
\begin{equation}
\label{eq:local_linearity_from_dc}
\mathrm{LL}_{a\rightarrow b}^{i}(\partition)
:=
\frac{1}{m}\sum_{r=1}^{m}\mathrm{DC}_{a\rightarrow b}^{i}(S_r).
\end{equation}

\begin{definition}[Cross-regime direction consistency]
\label{def:direction_consistency_cross}
This measures whether a direction estimated in one regime is preserved in another regime.
Let $\partition=\{S_r\}_{r=1}^m$ and fix source regime $S\in\partition$ and evaluation regime $S'\in\partition$. Define
\[
\hat v_{a\rightarrow b}^{S,i}:=\frac{1}{n_S}\sum_{j=1}^{n_S}\Delta_j^{S,i},
\]
Let $\mathcal{C}_{a\rightarrow b}^{S',i}=\{(z_t,z_t')\}_{t=1}^{n_{S'}}$ and $\Delta_t^{S',i}:=\phi(z_t')-\phi(z_t)$. The cross-regime DC score averages cosine alignment between evaluation displacements $\Delta_t^{S',i}$ and the source estimated concept direction $\hat v_{a\rightarrow b}^{S,i}$:
\begin{equation}
\label{eq:dc_cross}
\mathrm{DC}_{a\rightarrow b}^{i}(S,S'):=
\frac{1}{n_{S'}}\sum_{t=1}^{n_{S'}}
\cos\!\left(\Delta_t^{S',i},\hat v_{a\rightarrow b}^{S,i}\right).
 \end{equation}

High cross-regime DC across all regime pairs implies high global linearity.
We call $\mathrm{DC}_{a\rightarrow b}^{i}(S,S')$ \textbf{cross-regime DC}.
\end{definition}

We summarize global linearity by averaging cross-regime DC across distinct regime pairs:
\begin{equation}
\label{eq:global_linearity_from_dc}
\mathrm{GL}_{a\rightarrow b}^{i}(\partition)
:=
\frac{1}{m(m-1)}\sum_{r=1}^{m}\sum_{\substack{r'=1\\r'\neq r}}^{m}
\mathrm{DC}_{a\rightarrow b}^{i}(S_r,S_{r'}).
\end{equation}
Excluding identical regimes ($r'=r$) isolates global linearity from local linearity and allows us to understand whether the concept representation survives across regimes.

\subsubsection{Random baseline}
\label{sec:random_baseline}

DC scores (\cref{eq:dc_local,eq:dc_cross}) are informative of (local or global) linearity only if cosine alignment between evaluation displacements and the estimated concept direction $\hat v_{a\rightarrow b}^{S,i}$ exceeds chance level. As a baseline, we replace the evaluation pairs with pairs drawn uniformly at random from the pooled dataset for concept $i$ (e.g., all counting datasets for numerosity, all MIT-States images for MIT-States attributes), without the counterfactual matching of \cref{sec:counterfactuals}, and compute \cref{eq:dc_local,eq:dc_cross} with these random displacements.

To test whether a DC score is distinguishable from the baseline, we use a one-sided Welch $t$-test comparing per-regime-pair DC scores against the random baseline. For a given concept and model, we run this test for each of the $m^2$ regime pairs. Because the tests are conducted simultaneously, we apply Bonferroni correction to the error rate and require $p < \frac{0.05}{m^2}$. The random baseline along with the DC metric allows us to define fragmentation and characterize fragmented linearity.

\subsubsection{Fragmentation}
\label{sec:fragmentation}

We refer to a concept representation as \emph{fragmented} if no consistent concept direction transfers across regimes. We quantify this as the fraction of cross-regime pairs $(S_r, S_{r'})$, $r\neq r'$, where $\mathrm{DC}_{a\rightarrow b}^{i}(S_r,S_{r'})$ is indistinguishable from the random baseline.

\begin{definition}[Fragmentation score]
\label{def:fragmentation}
Fix concept $i$ and a partition $\partition=\{S_r\}_{r=1}^m$. Let $\mathcal{T}=\{(S_r,S_{r'})\mid r\neq r'\}$ be the $m(m-1)$ cross-regime pairs. For each source regime $S_r$, let $\mathrm{DC}_{a\rightarrow b}^{i,\mathrm{rand}}(S_r)$ be the random-baseline DC score obtained by replacing the evaluation pairs with uniformly random pairs from the pooled dataset (\cref{sec:random_baseline}). For each $(S_r,S_{r'})\in\mathcal{T}$, define the null hypothesis
\[
H_0^{r,r'}: \mathrm{DC}_{a\rightarrow b}^{i}(S_r,S_{r'}) \leq \mathrm{DC}_{a\rightarrow b}^{i,\mathrm{rand}}(S_r).
\]
Let $p_{r,r'}$ be the $p$-value from the one-sided Welch $t$-test of $H_0^{r,r'}$. The \textbf{fragmentation score} is
\begin{equation}
\label{eq:fragmentation}
\mathrm{Frag}_{a\rightarrow b}^i(\partition) := \frac{1}{m(m-1)}\sum_{r=1}^{m}\sum_{\substack{r'=1\\r'\neq r}}^{m} \mathbf{1}\!\left[p_{r,r'} \geq \frac{0.05}{m^2}\right].
\end{equation}
$\mathrm{Frag}_{a\rightarrow b}^i(\partition)=0$ indicates the concept representation is consistent across all regime pairs, while $\mathrm{Frag}_{a\rightarrow b}^i(\partition)=1$ indicates it is fragmented across all of them (under the sample sizes considered by us). 
\end{definition}

\begin{figure}[t]
    \centering
    \includegraphics[width=\linewidth]{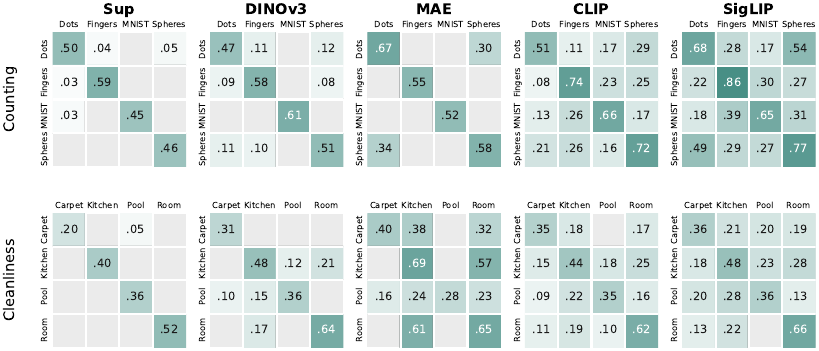}
    \caption{
    \textbf{Within-regime (diagonal) and cross-regime (off-diagonal) direction consistencies}. Some values are indistinguishable from baseline (according to Welch's t-test with Bonferroni-corrected p-values) and are excluded from display.
    Even supervised models (leftmost panel) are locally linear, while only selectively globally linear. In contrast, CLIP models are linear almost everywhere (most off-diagonal entries significant).
    }
    \label{fig:heatmaps_four}
\vspace{-1.5em}
\end{figure}

\subsection{Experimental setup}
\label{sec:experimental_setup}

We describe the setup for reproducing our experiments. 

\paragraph{Models.}

Our study covers \NUMMODELS\ vision models trained using different objectives: image-text contrastive models 
\cite{radford2021learning, zhai_sigmoid_2023, tschannen_siglip_2025}, 
DINO \cite{caron2021emergingpropertiesselfsupervised, oquab2024dinov2, siméoni2025dinov3}, 
masked image modeling \cite{he_masked_2021}, 
and several supervised models. We cover both CNN-based \cite{he2016deep} and transformer-based
\cite{dosovitskiy2021image} architectures.
For CNNs we use average-pooled features; for vision transformers we use the class token. 
Pretrained CLIP variants are sourced from OpenCLIP \cite{ilharco_gabriel_2021_5143773} and all other models from timm \cite{rw2019timm}. 
We provide the full list of models in \cref{app:models-section}.

\paragraph{Datasets.}

We consider \NUMCONCEPTFAMILIES\ concept families. 
For numerosity, we consider MNIST \cite{li_deng_mnist_2012}, Fingers \cite{koryakin_fingers}, 
CLEVR \cite{johnson2016clevr}, and a synthetic dot-counting dataset (100 images per regime-class pair; 4 regimes). 
For age, we consider UTKFace \cite{zhang2017age} and 11k-Hands \cite{afifi_11k_2018} (74 images per regime-class pair; 2 regimes). 
For object identity, we use PACS \cite{li_deeper_2017} ($\approx$151 images per regime-class pair; 4 regimes). 
For MIT-States \cite{isola2015discoveringstatestransformations}, we use 7 binary attributes (brightness, cleanliness, dryness, modernness, rawness, ripeness, sunniness), with 30 images per regime-class pair and 4-10 regimes per attribute.

\section{Results}

We evaluate \NUMMODELS\ vision models spanning contrastive, self-supervised, and supervised training objectives, on \NUMCONCEPTFAMILIES\ concepts (numerosity, age, object identity, and seven MIT-States attributes).
For each concept, images are partitioned into regimes (e.g., MNIST digits, raised fingers, rendered spheres for numerosity).
For each model-concept pair, we compute per-regime-pair $\mathrm{DC}$, yielding regime $\times$ regime matrices (\cref{fig:heatmaps_four}). Each entry in such a matrix is a measure of local linearity within that regime (diagonal entries) or global linearity across regimes (off-diagonal entries). 
Full details of models and datasets are in \cref{sec:experimental_setup}.

\paragraph{Vision embeddings exhibit locally linear structure.}
\cref{tab:aggregate-concept-metrics} (second column) shows that \LocalLinearityPct\% of all (model, concept, regime) triplets (over a total of \LocalLinearityTests\ tests) yield local DC scores statistically above baseline (per-matrix Bonferroni-corrected Welch's $t$-test), with the
lowest rate being \LocalLinearityFloorPctDryWet\% for dry$\rightarrow$wet.
In particular, from the representative results in \cref{fig:heatmaps_four}, diagonal entries (local linearities) in the DC matrices are consistently above zero across models and concepts.
Even for a supervised ViT (leftmost column), $\mathrm{DC}_{2\rightarrow 3}^{\mathrm{MNIST}} = \LocalDCViTSupMNISTCount$
and $\mathrm{DC}_{\mathrm{clean}\rightarrow\mathrm{dirty}}^{\mathrm{room}} = \LocalDCViTSupRoomCleanliness$, significant against near-zero random baselines.
We conclude that statistically detectable (but not perfect) local linearity is almost universal across models, concepts and regimes.

\begin{table}[t]
    \centering
    \small
    \caption{
    \textbf{Local linearity is detectable near-universally; global linearity is detectable much less often. Local and global linearities are positively correlated for all concepts.}
    We report fractions of (model, regime) pairs where within-regime DC and cross-regime DC are statistically distinguishable from baseline (per-matrix Bonferroni correction, $\alpha/m^2$), across 46 models.
    }
    \vspace{0.5em}
    \label{tab:aggregate-concept-metrics}
    \setlength{\tabcolsep}{4.5pt}
    \begin{tabular}{lcccc}
\toprule
\textbf{Concept} & \textbf{Local sig.\ (\%)} & \textbf{Global sig.\ (\%)} & \textbf{Kendall $\tau$} & \textbf{$p$} \\
\midrule
2 $\to$ 3 & 100.0 & 46.9 & 0.68 & $3.20 \times 10^{-11}$ \\
Clean $\to$ Dirty & 99.5 & 51.8 & 0.59 & $5.78 \times 10^{-9}$ \\
Dry $\to$ Wet & 97.8 & 18.3 & 0.52 & $3.02 \times 10^{-7}$ \\
Ancient $\to$ Modern & 100.0 & 81.7 & 0.57 & $3.04 \times 10^{-8}$ \\
Bright $\to$ Dark & 99.5 & 77.2 & 0.79 & $1.03 \times 10^{-14}$ \\
Cooked $\to$ Raw & 97.8 & 44.7 & 0.51 & $6.05 \times 10^{-7}$ \\
Ripe $\to$ Unripe & 100.0 & 68.5 & 0.38 & $2.30 \times 10^{-4}$ \\
Foggy $\to$ Sunny & 99.6 & 85.5 & 0.77 & $4.48 \times 10^{-14}$ \\
Dog $\to$ Horse & 100.0 & 99.3 & 0.89 & $2.78 \times 10^{-18}$ \\
Young $\to$ Old & 100.0 & 94.6 & 0.53 & $1.82 \times 10^{-7}$ \\
\midrule
\textit{Overall} & \textit{99.1} & \textit{52.2} &  &  \\
\bottomrule
\end{tabular}

    \vspace{-1em}
\end{table}

\paragraph{Linearity is often fragmented across regimes.}
From \cref{tab:aggregate-concept-metrics} (third column), only \GlobalLinearityPct\%
of all (model, concept, regime-pair) tuples yield cross-regime DC above baseline
(over \GlobalLinearityTests\ tests), compared to \LocalLinearityPct\% for local linearity. 
We show qualitative breakdowns across regime-pairs in \cref{fig:heatmaps_four}.
For the supervised ViT on numerosity ($2{\to}3$), in \FragInsigViTSupCounting\ of 12 regime pairs, cross-regime DC is indistinguishable from the random baseline (fragmentation $= \FragViTSupCounting$), despite significant local DCs. On the other hand, CLIP achieves fragmentation $= 0$ for the same transition: all cross-regime DCs are significant. Similarly, for cleanliness, supervised ViT has fragmentation $\FragViTSupCleanliness$ (\FragInsigViTSupCleanliness/12 cross-regime pairs near baseline), while CLIP has fragmentation $\FragCLIPCleanliness$ (only one insignificant pair).
We conclude that regime-specific linearity can coexist with near-absent global linearity.

\textbf{Global linearity does hold almost universally for object identity and age}, with (\GlobalLinearityPctObjects\%) and (\GlobalLinearityPctAge\%) of all statistical significance tests passing respectively.
Intuitively, for objects and age, regimes differ in style (photo vs.\ sketch, face vs.\ hand) while the visual cues for the concept (shape for objects, wrinkles for age) remain consistent. For more abstract concepts like numerosity and cleanliness, regimes differ in subject matter (digits, fingers, spheres), making it harder for regime-invariant directions to emerge.

Our findings above indicate that concept representations in vision embeddings can be significantly locally linear yet fully fragmented across regimes (\cref{fig:heatmaps_four}, left). 
While we have so far treated linearity as binary (either statistically significant or not) we now ask whether global linearity varies across models of different local linearities.
Does stronger within-regime linear structure in modern foundation models reflect deeper concept understanding, or simply stronger regime-specific biases (leading to even weaker global understanding)?

\paragraph{Global linearity increases with local linearity.}

\begin{figure}[t]
    \centering
    \includegraphics[width=0.91\linewidth]{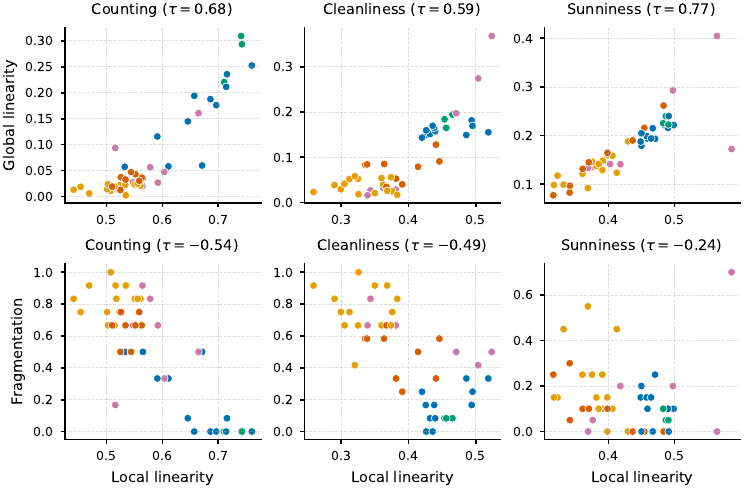}
    \caption{%
        \textbf{Top row:} Global linearity vs local linearity across models ({\footnotesize
        \textcolor[HTML]{0072B2}{$\bullet$}~CLIP\quad
        \textcolor[HTML]{009E73}{$\bullet$}~SigLIP\quad
        \textcolor[HTML]{D55E00}{$\bullet$}~DINO\quad
        \textcolor[HTML]{CC79A7}{$\bullet$}~MIM\quad
        \textcolor[HTML]{E69F00}{$\bullet$}~Sup}).
        \textbf{Bottom row:} Fragmentation vs local linearity.
        For numerosity ($2{\to}3$), MIT-States cleanliness (clean$\to$dirty), and sunniness (foggy$\to$sunny).}%
    \label{fig:trends_three}
    \vspace{-1em}
    \end{figure} 
From \cref{tab:aggregate-concept-metrics}, the rank correlation, measured by Kendall $\tau$, between global and local linearity across the tested models is positive for every concept ($\tau \in [0.38, 0.89]$, all $p < 10^{-3}$). 
As qualitative examples, \cref{fig:trends_three} plots global linearity (\cref{eq:global_linearity_from_dc}) and fragmentation (\cref{eq:fragmentation}) against local linearity (\cref{eq:local_linearity_from_dc}) across all models, for representative concept transitions (2$\to$3, clean$\to$dirty, foggy$\to$sunny). Global linearity correlates positively with local linearity in all three cases ($\tau = 0.68$, $0.59$, $0.77$). 
This finding echoes the platonic representation hypothesis~\citep{huh_platonic_2024}: stronger local concept structure in vision models converges toward abstract, regime-invariant geometry. 
Crucially, we do not claim a causal effect; our result describes the trend that is emerging in currently popular vision foundation models. 

\paragraph{Concept directions can act as cross-regime linear probes in globally linear models.}

Regime-invariant geometry (indicated by high global linearity) implies that concept directions should transfer as linear probes across regimes.
For source regime $S$ and target regime $S'\neq S$, we classify held-out samples from $S'$ using
the direction $\hat{v}_{a\rightarrow b}^{S,i}$ estimated in $S$ alone: sample $z\in S'$ is assigned
class $b$ if $\bigl(\phi(z)-\hat{\mu}_{S'}\bigr)\cdot\hat{v}_{a\rightarrow b}^{S,i}>0$,
where $\hat{\mu}_{S'}$ is the mean embedding of $S'$ (computed using 5 randomly selected samples per class) and provides an estimate of the mid-point between the two classes.

\begin{figure}[tp]
    \centering
    \includegraphics[width=0.85\linewidth]{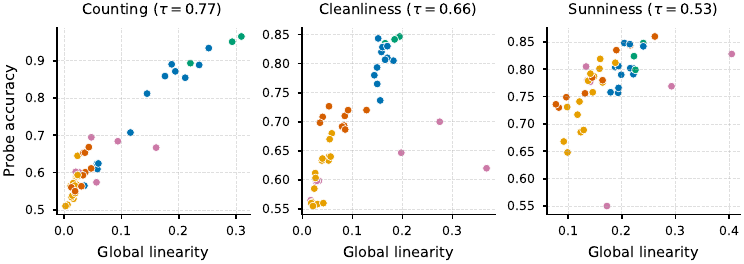}
    \caption{%
        Probe accuracy vs global linearity across models ({\footnotesize
        \textcolor[HTML]{0072B2}{$\bullet$}~CLIP\quad
        \textcolor[HTML]{009E73}{$\bullet$}~SigLIP\quad
        \textcolor[HTML]{D55E00}{$\bullet$}~DINO\quad
        \textcolor[HTML]{CC79A7}{$\bullet$}~MIM\quad
        \textcolor[HTML]{E69F00}{$\bullet$}~Sup})
        for numerosity ($2{\to}3$), cleanliness (clean$\to$dirty), and sunniness (foggy$\to$sunny).}%
    \label{fig:probe_trends_three}
    \vspace{-1.5em}
\end{figure}

\cref{fig:probe_trends_three} plots probe accuracy against global linearity $\mathrm{GL}_{a\rightarrow b}^{i}(\partition)$ across models for three representative transitions.
Probe accuracy correlates positively with global linearity in all three cases ($\tau = \ProbeTauCount$, $\ProbeTauCleanliness$, $\ProbeTauSunny$).
We conclude that concept directions in high-global-linearity models are indeed semantically meaningful.

\section{Conclusion}

Almost all vision models encode \quotes{three}, but not all of them know it is \quotes{three}. 
While vision embeddings are locally linear to the extent of statistical significance almost without exception, many of them learn the same concept several times, once per regime, reflected in the phenomenon we characterize as fragmentation. As locally linear structure gets stronger, cross-regime alignment increases and global linearity emerges, producing semantically meaningful concept directions that can act as effective linear probes. Given our characterization of vision embedding geometry, the overarching question as to whether regime-specific \emph{skills} are also converging into abstract ones (and whether they are related to geometry) remains open to future work.

\paragraph{Limitations and broader impact.}

Although we cover a broad range of concepts exhibiting varying degrees of difficulty, our conclusions are limited in scope to the investigated concept types. Our counterfactual pairs are approximate: matched by embedding space distance. While these limitations are unavoidable in the current scope, we argue that our statistical tests and downstream capability (probing) results together establish sufficient evidence to validate our main claims. 

Linear directions in embedding space can potentially be exploited to steer representations toward harmful outputs; our work characterizes this geometry but does not introduce new attack surfaces beyond what already exists in existing models. On the other hand, our work contributes to the literature on interpretability, which strives to make AI systems fair and transparent.

\paragraph{Acknowledgements.}
This work was partly supported by the National Research Foundation of Korea (NRF) grant funded by the Korean government (MSIT) in 2026 (Grant No. RS-2026-25612350, Project Title: Development of an End-to-End Design and Validation Framework for Mathematically Principled Next-Generation AI Architectures and Learning Algorithms). This work was partly also supported by Institute for Information \& communications Technology Planning \& Evaluation (IITP) grant funded by the Korea government (MSIT) (RS-2019-II190075, Artificial Intelligence Graduate School Program(KAIST)).
The authors are grateful to Nishad Singhi and Davide Berasi for helpful feedback.
AS and AU acknowledge support from the German Federal Ministry of Education and Research (BMBF): Tübingen AI Center, FKZ: 01IS18039A, and the International Max Planck Research School for Intelligent Systems (IMPRS-IS).

\newpage
\bibliographystyle{unsrtnat}
\bibliography{references,references_manual}

@misc{kumar_questioning_2025,
    title = {Questioning {Representational} {Optimism} in {Deep} {Learning}: {The} {Fractured} {Entangled} {Representation} {Hypothesis}},
    shorttitle = {Questioning {Representational} {Optimism} in {Deep} {Learning}},
    url = {http://arxiv.org/abs/2505.11581},
    doi = {10.48550/arXiv.2505.11581},
    urldate = {2025-08-22},
    publisher = {arXiv},
    author = {Kumar, Akarsh and Clune, Jeff and Lehman, Joel and Stanley, Kenneth O.},
    month = may,
    year = {2025},
    note = {arXiv:2505.11581 [cs]},
}

@misc{lepori_break_2023,
    title = {Break {It} {Down}: {Evidence} for {Structural} {Compositionality} in {Neural} {Networks}},
    shorttitle = {Break {It} {Down}},
    url = {http://arxiv.org/abs/2301.10884},
    doi = {10.48550/arXiv.2301.10884},
    urldate = {2025-09-04},
    publisher = {arXiv},
    author = {Lepori, Michael A. and Serre, Thomas and Pavlick, Ellie},
    month = nov,
    year = {2023},
    note = {arXiv:2301.10884 [cs]},
}

@article{huh_platonic_2024,
    title = {The {Platonic} {Representation} {Hypothesis}},
    url = {http://arxiv.org/abs/2405.07987},
    language = {en},
    urldate = {2024-07-05},
    journal = {ICML},
    author = {Huh, Minyoung and Cheung, Brian and Wang, Tongzhou and Isola, Phillip},
    month = may,
    year = {2024},
    note = {arXiv:2405.07987 [cs]},
}

@article{thasarathan_universal_2025,
    title = {Universal {Sparse} {Autoencoders}: {Interpretable} {Cross}-{Model} {Concept} {Alignment}},
    language = {en},
    author = {Thasarathan, Harrish and Forsyth, Julian and Fel, Thomas and Kowal, Matthew and Derpanis, Konstantinos G},
    month = feb,
    year = {2025},
}

@article{al-tahan_scaling_nodate,
    title = {Scaling {Vision}-{Language} {Models} {Does} {Not} {Improve} {Relational} {Understanding}: {The} {Right} {Learning} {Objective} {Helps}},
    language = {en},
    author = {Al-Tahan, Haider and Garrido, Quentin and Balestriero, Randall and Bouchacourt, Diane and Hazirbas, Caner and Ibrahim, Mark},
}

@inproceedings{liang_mind_2022,
    title = {Mind the {Gap}: {Understanding} the {Modality} {Gap} in {Multi}-modal {Contrastive} {Representation} {Learning}},
    shorttitle = {Mind the {Gap}},
    url = {https://openreview.net/forum?id=S7Evzt9uit3},
    language = {en},
    urldate = {2025-09-29},
    author = {Liang, Weixin and Zhang, Yuhui and Kwon, Yongchan and Yeung, Serena and Zou, James},
    month = oct,
    year = {2022},
}

@article{schrodi_two_2025,
    title = {Two {Effects}, {One} {Trigger}: {On} the {Modality} {Gap}, {Object} {Bias}, and {Information} {Imbalance} in {Contrastive} {Vision}-{Language} {Models}},
    language = {en},
    author = {Schrodi, Simon and Hoffmann, David T and Argus, Max and Fischer, Volker and Brox, Thomas},
    year = {2025},
}

@misc{trager_linear_2024,
    title = {Linear {Spaces} of {Meanings}: {Compositional} {Structures} in {Vision}-{Language} {Models}},
    shorttitle = {Linear {Spaces} of {Meanings}},
    url = {http://arxiv.org/abs/2302.14383},
    doi = {10.48550/arXiv.2302.14383},
    urldate = {2025-07-11},
    publisher = {arXiv},
    author = {Trager, Matthew and Perera, Pramuditha and Zancato, Luca and Achille, Alessandro and Bhatia, Parminder and Soatto, Stefano},
    month = jan,
    year = {2024},
    note = {arXiv:2302.14383 [cs]},
}

@misc{kang_is_2025,
    title = {Is {CLIP} ideal? {No}. {Can} we fix it? {Yes}!},
    shorttitle = {Is {CLIP} ideal?},
    url = {http://arxiv.org/abs/2503.08723},
    doi = {10.48550/arXiv.2503.08723},
    urldate = {2025-12-22},
    publisher = {arXiv},
    author = {Kang, Raphi and Song, Yue and Gkioxari, Georgia and Perona, Pietro},
    month = mar,
    year = {2025},
    note = {arXiv:2503.08723 [cs]},
}

@misc{ranjbar_spacevlm_2025,
    title = {{SpaceVLM}: {Sub}-{Space} {Modeling} of {Negation} in {Vision}-{Language} {Models}},
    shorttitle = {{SpaceVLM}},
    url = {http://arxiv.org/abs/2511.12331},
    doi = {10.48550/arXiv.2511.12331},
    urldate = {2025-12-22},
    publisher = {arXiv},
    author = {Ranjbar, Sepehr Kazemi and Alhamoud, Kumail and Ghassemi, Marzyeh},
    month = nov,
    year = {2025},
    note = {arXiv:2511.12331 [cs]},
}

@misc{berasi_not_2025,
    title = {Not {Only} {Text}: {Exploring} {Compositionality} of {Visual} {Representations} in {Vision}-{Language} {Models}},
    shorttitle = {Not {Only} {Text}},
    url = {http://arxiv.org/abs/2503.17142},
    doi = {10.48550/arXiv.2503.17142},
    urldate = {2025-12-22},
    publisher = {arXiv},
    author = {Berasi, Davide and Farina, Matteo and Mancini, Massimiliano and Ricci, Elisa and Strisciuglio, Nicola},
    month = mar,
    year = {2025},
    note = {arXiv:2503.17142 [cs]},
}

@inproceedings{radford2021learning,
    title = {Learning {Transferable} {Visual} {Models} {From} {Natural} {Language} {Supervision}},
    url = {https://proceedings.mlr.press/v139/radford21a.html},
    language = {english},
    urldate = {2024-07-01},
    booktitle = {Proceedings of the 38th {International} {Conference} on {Machine} {Learning}},
    publisher = {PMLR},
    author = {Radford, Alec and Kim, Jong Wook and Hallacy, Chris and Ramesh, Aditya and Goh, Gabriel and Agarwal, Sandhini and Sastry, Girish and Askell, Amanda and Mishkin, Pamela and Clark, Jack and Krueger, Gretchen and Sutskever, Ilya},
    month = jul,
    year = {2021},
    pages = {8748--8763},
}

@article{sonthalia_rankability_2025,
    title = {On the rankability of visual embeddings},
    url = {http://arxiv.org/abs/2507.03683},
    doi = {10.48550/arXiv.2507.03683},
    urldate = {2025-07-11},
    journal = {NeurIPS},
    author = {Sonthalia, Ankit and Uselis, Arnas and Oh, Seong Joon},
    year = {2025},
    note = {arXiv:2507.03683 [cs]},
}

@inproceedings{wolff_independent_2023,
    title = {The {Independent} {Compositional} {Subspace} {Hypothesis} for the {Structure} of {CLIP}'s {Last} {Layer}},
    url = {https://openreview.net/forum?id=MmhGK8YkUKO},
    language = {en},
    urldate = {2026-01-14},
    author = {Wolff, Max and Brendel, Wieland and Wolff, Stuart},
    month = feb,
    year = {2023},
}

@article{qu_learning_2025,
    title = {{LEARNING} {CLUSTERING}-{BASED} {PROTOTYPES} {FOR} {COMPOSITIONAL} {ZERO}-{SHOT} {LEARNING}},
    language = {en},
    author = {Qu, Hongyu and Wei, Jianan and Shu, Xiangbo and Wang, Wenguan},
    year = {2025},
}

@misc{bhalla_interpreting_2024,
    title = {Interpreting {CLIP} with {Sparse} {Linear} {Concept} {Embeddings} ({SpLiCE})},
    url = {http://arxiv.org/abs/2402.10376},
    urldate = {2024-10-16},
    publisher = {arXiv},
    author = {Bhalla, Usha and Oesterling, Alex and Srinivas, Suraj and Calmon, Flavio P. and Lakkaraju, Himabindu},
    month = feb,
    year = {2024},
    note = {arXiv:2402.10376},
}

@misc{park2024linear,
    title = {The {Linear} {Representation} {Hypothesis} and the {Geometry} of {Large} {Language} {Models}},
    url = {http://arxiv.org/abs/2311.03658},
    urldate = {2025-01-05},
    author = {Park, Kiho and Choe, Yo Joong and Veitch, Victor},
    month = jul,
    year = {2024},
    doi = {10.48550/arXiv.2311.03658},
    note = {arXiv: 2311.03658 [cs]},
}

@misc{liu_visual_2023,
    title = {Visual {Instruction} {Tuning}},
    url = {http://arxiv.org/abs/2304.08485},
    doi = {10.48550/arXiv.2304.08485},
    urldate = {2026-01-19},
    publisher = {arXiv},
    author = {Liu, Haotian and Li, Chunyuan and Wu, Qingyang and Lee, Yong Jae},
    month = dec,
    year = {2023},
    note = {arXiv:2304.08485 [cs]},
}

@article{li_deng_mnist_2012,
    title = {The {MNIST} {Database} of {Handwritten} {Digit} {Images} for {Machine} {Learning} {Research} [{Best} of the {Web}]},
    volume = {29},
    copyright = {https://ieeexplore.ieee.org/Xplorehelp/downloads/license-information/IEEE.html},
    url = {http://ieeexplore.ieee.org/document/6296535/},
    doi = {10.1109/MSP.2012.2211477},
    language = {en},
    number = {6},
    urldate = {2026-01-19},
    journal = {IEEE Signal Processing Magazine},
    author = {{Li Deng}},
    month = nov,
    year = {2012},
    pages = {141--142},
}

@misc{johnson2016clevr,
    title = {{CLEVR}: {A} {Diagnostic} {Dataset} for {Compositional} {Language} and {Elementary} {Visual} {Reasoning}},
    shorttitle = {{CLEVR}},
    url = {http://arxiv.org/abs/1612.06890},
    urldate = {2024-12-20},
    author = {Johnson, Justin and Hariharan, Bharath and van der Maaten, Laurens and Fei-Fei, Li and Zitnick, C. Lawrence and Girshick, Ross},
    month = dec,
    year = {2016},
    doi = {10.48550/arXiv.1612.06890},
    note = {arXiv: 1612.06890 [cs]},
}

@inproceedings{kim_interpretability_2018,
    title = {Interpretability {Beyond} {Feature} {Attribution}: {Quantitative} {Testing} with {Concept} {Activation} {Vectors} ({TCAV})},
    shorttitle = {Interpretability {Beyond} {Feature} {Attribution}},
    url = {https://proceedings.mlr.press/v80/kim18d.html},
    language = {en},
    urldate = {2026-01-19},
    booktitle = {Proceedings of the 35th {International} {Conference} on {Machine} {Learning}},
    publisher = {PMLR},
    author = {Kim, Been and Wattenberg, Martin and Gilmer, Justin and Cai, Carrie and Wexler, James and Viegas, Fernanda and Sayres, Rory},
    month = jul,
    year = {2018},
    pages = {2668--2677},
}

@inproceedings{yeh_completeness-aware_2020,
    title = {On {Completeness}-aware {Concept}-{Based} {Explanations} in {Deep} {Neural} {Networks}},
    volume = {33},
    url = {https://proceedings.neurips.cc/paper/2020/hash/ecb287ff763c169694f682af52c1f309-Abstract.html},
    urldate = {2026-01-19},
    booktitle = {Advances in {Neural} {Information} {Processing} {Systems}},
    publisher = {Curran Associates, Inc.},
    author = {Yeh, Chih-Kuan and Kim, Been and Arik, Sercan and Li, Chun-Liang and Pfister, Tomas and Ravikumar, Pradeep},
    year = {2020},
    pages = {20554--20565},
}

@misc{alain_understanding_2018,
    title = {Understanding intermediate layers using linear classifier probes},
    url = {http://arxiv.org/abs/1610.01644},
    doi = {10.48550/arXiv.1610.01644},
    urldate = {2026-01-19},
    publisher = {arXiv},
    author = {Alain, Guillaume and Bengio, Yoshua},
    month = nov,
    year = {2018},
    note = {arXiv:1610.01644 [stat]},
}

@misc{uselis_does_2025,
    title = {Does {Data} {Scaling} {Lead} to {Visual} {Compositional} {Generalization}?},
    url = {http://arxiv.org/abs/2507.07102},
    doi = {10.48550/arXiv.2507.07102},
    urldate = {2026-01-20},
    publisher = {arXiv},
    author = {Uselis, Arnas and Dittadi, Andrea and Oh, Seong Joon},
    month = jul,
    year = {2025},
    note = {arXiv:2507.07102 [cs]},
}

@misc{uselis_intermediate_2025,
    title = {Intermediate {Layer} {Classifiers} for {OOD} generalization},
    url = {http://arxiv.org/abs/2504.05461},
    doi = {10.48550/arXiv.2504.05461},
    urldate = {2026-01-21},
    publisher = {arXiv},
    author = {Uselis, Arnas and Oh, Seong Joon},
    month = apr,
    year = {2025},
    note = {arXiv:2504.05461 [cs]},
}

@inproceedings{mikolov2013linguisticregularitiescontinuous,
    address = {Atlanta, Georgia},
    title = {Linguistic {Regularities} in {Continuous} {Space} {Word} {Representations}},
    url = {https://aclanthology.org/N13-1090/},
    urldate = {2026-01-19},
    booktitle = {Proceedings of the 2013 {Conference} of the {North} {American} {Chapter} of the {Association} for {Computational} {Linguistics}: {Human} {Language} {Technologies}},
    publisher = {Association for Computational Linguistics},
    author = {Mikolov, Tomas and Yih, Wen-tau and Zweig, Geoffrey},
    editor = {Vanderwende, Lucy and Daum{\'e} III, Hal and Kirchhoff, Katrin},
    month = jun,
    year = {2013},
    pages = {746--751},
}

@inproceedings{isola2015discoveringstatestransformations,
    address = {Boston, MA, USA},
    title = {Discovering states and transformations in image collections},
    isbn = {978-1-4673-6964-0},
    url = {http://ieeexplore.ieee.org/document/7298744/},
    doi = {10.1109/CVPR.2015.7298744},
    language = {en},
    urldate = {2025-01-05},
    booktitle = {2015 {IEEE} {Conference} on {Computer} {Vision} and {Pattern} {Recognition} ({CVPR})},
    publisher = {IEEE},
    author = {Isola, Phillip and Lim, Joseph J. and Adelson, Edward H.},
    month = jun,
    year = {2015},
    pages = {1383--1391},
}

@misc{chen2020simpleframeworkcontrastive,
    title = {A {Simple} {Framework} for {Contrastive} {Learning} of {Visual} {Representations}},
    url = {http://arxiv.org/abs/2002.05709},
    doi = {10.48550/arXiv.2002.05709},
    urldate = {2026-01-26},
    publisher = {arXiv},
    author = {Chen, Ting and Kornblith, Simon and Norouzi, Mohammad and Hinton, Geoffrey},
    month = jul,
    year = {2020},
    note = {arXiv:2002.05709 [cs]},
}

@misc{he2020momentumcontrastunsupervised,
    title = {Momentum {Contrast} for {Unsupervised} {Visual} {Representation} {Learning}},
    url = {http://arxiv.org/abs/1911.05722},
    doi = {10.48550/arXiv.1911.05722},
    urldate = {2026-01-26},
    publisher = {arXiv},
    author = {He, Kaiming and Fan, Haoqi and Wu, Yuxin and Xie, Saining and Girshick, Ross},
    month = mar,
    year = {2020},
    note = {arXiv:1911.05722 [cs]},
}

@misc{vompa2025linearseparabilityceiling,
    title = {Beyond the {Linear} {Separability} {Ceiling}: {Aligning} {Representations} in {VLMs}},
    shorttitle = {Beyond the {Linear} {Separability} {Ceiling}},
    url = {http://arxiv.org/abs/2507.07574},
    doi = {10.48550/arXiv.2507.07574},
    urldate = {2026-01-26},
    publisher = {arXiv},
    author = {Vompa, Enrico and Tammet, Tanel and Vaishnav, Mohit},
    month = dec,
    year = {2025},
    note = {arXiv:2507.07574 [cs]},
}

@misc{siméoni2025dinov3,
    title = {{DINOv3}},
    copyright = {arXiv.org perpetual, non-exclusive license},
    url = {https://arxiv.org/abs/2508.10104},
    doi = {10.48550/ARXIV.2508.10104},
    urldate = {2026-05-04},
    publisher = {arXiv},
    author = {Siméoni, Oriane and Vo, Huy V. and Seitzer, Maximilian and Baldassarre, Federico and Oquab, Maxime and Jose, Cijo and Khalidov, Vasil and Szafraniec, Marc and Yi, Seungeun and Ramamonjisoa, Michaël and Massa, Francisco and Haziza, Daniel and Wehrstedt, Luca and Wang, Jianyuan and Darcet, Timothée and Moutakanni, Théo and Sentana, Leonel and Roberts, Claire and Vedaldi, Andrea and Tolan, Jamie and Brandt, John and Couprie, Camille and Mairal, Julien and Jégou, Hervé and Labatut, Patrick and Bojanowski, Piotr},
    year = {2025},
    note = {Version Number: 1},
}

@inproceedings{caron2021emergingpropertiesselfsupervised,
    address = {Montreal, QC, Canada},
    title = {Emerging {Properties} in {Self}-{Supervised} {Vision} {Transformers}},
    copyright = {https://doi.org/10.15223/policy-029},
    isbn = {978-1-6654-2812-5},
    url = {https://ieeexplore.ieee.org/document/9709990/},
    doi = {10.1109/ICCV48922.2021.00951},
    language = {en},
    urldate = {2024-06-26},
    booktitle = {2021 {IEEE}/{CVF} {International} {Conference} on {Computer} {Vision} ({ICCV})},
    publisher = {IEEE},
    author = {Caron, Mathilde and Touvron, Hugo and Misra, Ishan and Jegou, Herve and Mairal, Julien and Bojanowski, Piotr and Joulin, Armand},
    month = oct,
    year = {2021},
    pages = {9630--9640},
}

@misc{rw2019timm,
  author = {Ross Wightman},
  title = {PyTorch Image Models},
  year = {2019},
  publisher = {GitHub},
  journal = {GitHub repository},
  doi = {10.5281/zenodo.4414861},
  howpublished = {\url{https://github.com/rwightman/pytorch-image-models}}
}

@software{ilharco_gabriel_2021_5143773,
  author       = {Ilharco, Gabriel and
                  Wortsman, Mitchell and
                  Wightman, Ross and
                  Gordon, Cade and
                  Carlini, Nicholas and
                  Taori, Rohan and
                  Dave, Achal and
                  Shankar, Vaishaal and
                  Namkoong, Hongseok and
                  Miller, John and
                  Hajishirzi, Hannaneh and
                  Farhadi, Ali and
                  Schmidt, Ludwig},
  title        = {OpenCLIP},
  month        = jul,
  year         = 2021,
  note         = {If you use this software, please cite it as below.},
  publisher    = {Zenodo},
  version      = {0.1},
  doi          = {10.5281/zenodo.5143773},
  url          = {https://doi.org/10.5281/zenodo.5143773}
}

@misc{koryakin_fingers,
  author       = {Koryakin, Pavel},
  title        = {Fingers Dataset},
  year         = {2019},
  howpublished = {\url{https://www.kaggle.com/datasets/koryakinp/fingers}},
  note         = {CC0: Public Domain}
}

@misc{oquab2024dinov2,
    title = {{DINOv2}: {Learning} {Robust} {Visual} {Features} without {Supervision}},
    shorttitle = {{DINOv2}},
    url = {http://arxiv.org/abs/2304.07193},
    urldate = {2025-05-14},
    author = {Oquab, Maxime and Darcet, Timothée and Moutakanni, Théo and Vo, Huy and Szafraniec, Marc and Khalidov, Vasil and Fernandez, Pierre and Haziza, Daniel and Massa, Francisco and El-Nouby, Alaaeldin and Assran, Mahmoud and Ballas, Nicolas and Galuba, Wojciech and Howes, Russell and Huang, Po-Yao and Li, Shang-Wen and Misra, Ishan and Rabbat, Michael and Sharma, Vasu and Synnaeve, Gabriel and Xu, Hu and Jegou, Hervé and Mairal, Julien and Labatut, Patrick and Joulin, Armand and Bojanowski, Piotr},
    month = feb,
    year = {2024},
    doi = {10.48550/arXiv.2304.07193},
    note = {arXiv: 2304.07193 [cs]},
}

@misc{fel_archetypal_2025,
    title = {Archetypal {SAE}: {Adaptive} and {Stable} {Dictionary} {Learning} for {Concept} {Extraction} in {Large} {Vision} {Models}},
    shorttitle = {Archetypal {SAE}},
    url = {http://arxiv.org/abs/2502.12892},
    doi = {10.48550/arXiv.2502.12892},
    urldate = {2026-01-26},
    publisher = {arXiv},
    author = {Fel, Thomas and Lubana, Ekdeep Singh and Prince, Jacob S. and Kowal, Matthew and Boutin, Victor and Papadimitriou, Isabel and Wang, Binxu and Wattenberg, Martin and Ba, Demba and Konkle, Talia},
    month = may,
    year = {2025},
    note = {arXiv:2502.12892 [cs]},
}

@misc{gorton_missing_2024,
    title = {The {Missing} {Curve} {Detectors} of {InceptionV1}: {Applying} {Sparse} {Autoencoders} to {InceptionV1} {Early} {Vision}},
    shorttitle = {The {Missing} {Curve} {Detectors} of {InceptionV1}},
    url = {http://arxiv.org/abs/2406.03662},
    doi = {10.48550/arXiv.2406.03662},
    urldate = {2026-01-26},
    publisher = {arXiv},
    author = {Gorton, Liv},
    month = sep,
    year = {2024},
    note = {arXiv:2406.03662 [cs]},
}

@misc{zhang2017age,
    title = {Age {Progression}/{Regression} by {Conditional} {Adversarial} {Autoencoder}},
    url = {http://arxiv.org/abs/1702.08423},
    urldate = {2025-03-22},
    author = {Zhang, Zhifei and Song, Yang and Qi, Hairong},
    month = mar,
    year = {2017},
    doi = {10.48550/arXiv.1702.08423},
    note = {arXiv: 1702.08423 [cs]},
}

@misc{afifi_11k_2018,
    title = {{11K} {Hands}: {Gender} recognition and biometric identification using a large dataset of hand images},
    shorttitle = {{11K} {Hands}},
    url = {http://arxiv.org/abs/1711.04322},
    doi = {10.48550/arXiv.1711.04322},
    urldate = {2026-01-26},
    publisher = {arXiv},
    author = {Afifi, Mahmoud},
    month = sep,
    year = {2018},
    note = {arXiv:1711.04322 [cs]},
}

@misc{li_deeper_2017,
    title = {Deeper, {Broader} and {Artier} {Domain} {Generalization}},
    url = {http://arxiv.org/abs/1710.03077},
    doi = {10.48550/arXiv.1710.03077},
    urldate = {2026-01-26},
    publisher = {arXiv},
    author = {Li, Da and Yang, Yongxin and Song, Yi-Zhe and Hospedales, Timothy M.},
    month = oct,
    year = {2017},
    note = {arXiv:1710.03077 [cs]},
}

@misc{zhai_sigmoid_2023,
    title = {Sigmoid {Loss} for {Language} {Image} {Pre}-{Training}},
    url = {http://arxiv.org/abs/2303.15343},
    doi = {10.48550/arXiv.2303.15343},
    urldate = {2026-01-26},
    publisher = {arXiv},
    author = {Zhai, Xiaohua and Mustafa, Basil and Kolesnikov, Alexander and Beyer, Lucas},
    month = sep,
    year = {2023},
    note = {arXiv:2303.15343 [cs]},
}

@misc{tschannen_siglip_2025,
    title = {{SigLIP} 2: {Multilingual} {Vision}-{Language} {Encoders} with {Improved} {Semantic} {Understanding}, {Localization}, and {Dense} {Features}},
    shorttitle = {{SigLIP} 2},
    url = {http://arxiv.org/abs/2502.14786},
    doi = {10.48550/arXiv.2502.14786},
    urldate = {2026-01-26},
    publisher = {arXiv},
    author = {Tschannen, Michael and Gritsenko, Alexey and Wang, Xiao and Naeem, Muhammad Ferjad and Alabdulmohsin, Ibrahim and Parthasarathy, Nikhil and Evans, Talfan and Beyer, Lucas and Xia, Ye and Mustafa, Basil and Hénaff, Olivier and Harmsen, Jeremiah and Steiner, Andreas and Zhai, Xiaohua},
    month = feb,
    year = {2025},
    note = {arXiv:2502.14786 [cs]},
}

@misc{he_masked_2021,
    title = {Masked {Autoencoders} {Are} {Scalable} {Vision} {Learners}},
    url = {http://arxiv.org/abs/2111.06377},
    urldate = {2024-02-17},
    publisher = {arXiv},
    author = {He, Kaiming and Chen, Xinlei and Xie, Saining and Li, Yanghao and Dollár, Piotr and Girshick, Ross},
    month = dec,
    year = {2021},
    note = {arXiv:2111.06377 [cs]},
}

@inproceedings{he2016deep,
    title = {Deep {Residual} {Learning} for {Image} {Recognition}},
    url = {http://ieeexplore.ieee.org/document/7780459/},
    doi = {10.1109/CVPR.2016.90},
    language = {english},
    urldate = {2022-11-30},
    booktitle = {2016 {IEEE} {Conference} on {Computer} {Vision} and {Pattern} {Recognition} ({CVPR})},
    publisher = {IEEE},
    author = {He, Kaiming and Zhang, Xiangyu and Ren, Shaoqing and Sun, Jian},
    month = jun,
    year = {2016},
    note = {Place: Las Vegas, NV, USA},
    pages = {770--778},
}

@misc{dosovitskiy2021image,
    title = {An {Image} {Is} {Worth} 16x16 {Words}: {Transformers} for {Image} {Recognition} at {Scale}},
    shorttitle = {An {Image} {Is} {Worth} 16x16 {Words}},
    url = {http://arxiv.org/abs/2010.11929},
    urldate = {2024-11-23},
    author = {Dosovitskiy, Alexey and Beyer, Lucas and Kolesnikov, Alexander and Weissenborn, Dirk and Zhai, Xiaohua and Unterthiner, Thomas and Dehghani, Mostafa and Minderer, Matthias and Heigold, Georg and Gelly, Sylvain and Uszkoreit, Jakob and Houlsby, Neil},
    month = jun,
    year = {2021},
    doi = {10.48550/arXiv.2010.11929},
    note = {arXiv: 2010.11929},
}

@misc{levi2024doubleellipsoid,
    title = {The {Double}-{Ellipsoid} {Geometry} of {CLIP}},
    url = {http://arxiv.org/abs/2411.14517},
    urldate = {2025-05-01},
    author = {Levi, Meir Yossef and Gilboa, Guy},
    month = nov,
    year = {2024},
    doi = {10.48550/arXiv.2411.14517},
    note = {arXiv: 2411.14517 [cs]},
}

@misc{koishigarina2025clip,
    title = {{CLIP} {Behaves} like a {Bag}-of-{Words} {Model} {Cross}-modally but not {Uni}-modally},
    url = {http://arxiv.org/abs/2502.03566},
    urldate = {2025-02-13},
    author = {Koishigarina, Darina and Uselis, Arnas and Oh, Seong Joon},
    month = feb,
    year = {2025},
    doi = {10.48550/arXiv.2502.03566},
    note = {arXiv: 2502.03566 [cs]},
}

@misc{lewis_does_2024,
    title = {Does {CLIP} {Bind} {Concepts}? {Probing} {Compositionality} in {Large} {Image} {Models}},
    shorttitle = {Does {CLIP} {Bind} {Concepts}?},
    url = {http://arxiv.org/abs/2212.10537},
    doi = {10.48550/arXiv.2212.10537},
    urldate = {2026-01-28},
    publisher = {arXiv},
    author = {Lewis, Martha and Nayak, Nihal V. and Yu, Peilin and Yu, Qinan and Merullo, Jack and Bach, Stephen H. and Pavlick, Ellie},
    month = aug,
    year = {2024},
    note = {arXiv:2212.10537 [cs]},
}

@misc{mccracken_uncovering_2025,
    title = {Uncovering a {Universal} {Abstract} {Algorithm} for {Modular} {Addition} in {Neural} {Networks}},
    url = {http://arxiv.org/abs/2505.18266},
    doi = {10.48550/arXiv.2505.18266},
    urldate = {2025-08-22},
    publisher = {arXiv},
    author = {McCracken, Gavin and Moisescu-Pareja, Gabriela and Letourneau, Vincent and Precup, Doina and Love, Jonathan},
    month = may,
    year = {2025},
    note = {arXiv:2505.18266 [cs]},
}

@inproceedings{nayak_learning_2022,
    title = {Learning to {Compose} {Soft} {Prompts} for {Compositional} {Zero}-{Shot} {Learning}},
    url = {https://openreview.net/forum?id=S8-A2FXnIh},
    language = {en},
    urldate = {2026-01-28},
    author = {Nayak, Nihal V. and Yu, Peilin and Bach, Stephen},
    month = sep,
    year = {2022},
}

@misc{perera_prompt_2023,
    title = {Prompt {Algebra} for {Task} {Composition}},
    url = {http://arxiv.org/abs/2306.00310},
    doi = {10.48550/arXiv.2306.00310},
    urldate = {2026-01-28},
    publisher = {arXiv},
    author = {Perera, Pramuditha and Trager, Matthew and Zancato, Luca and Achille, Alessandro and Soatto, Stefano},
    month = jun,
    year = {2023},
    note = {arXiv:2306.00310 [cs]},
}

@misc{baldrati_zero-shot_2023,
    title = {Zero-{Shot} {Composed} {Image} {Retrieval} with {Textual} {Inversion}},
    url = {http://arxiv.org/abs/2303.15247},
    doi = {10.48550/arXiv.2303.15247},
    urldate = {2026-01-29},
    publisher = {arXiv},
    author = {Baldrati, Alberto and Agnolucci, Lorenzo and Bertini, Marco and Bimbo, Alberto Del},
    month = aug,
    year = {2023},
    note = {arXiv:2303.15247 [cs]},
}

@misc{karthik_vision-by-language_2024,
    title = {Vision-by-{Language} for {Training}-{Free} {Compositional} {Image} {Retrieval}},
    url = {http://arxiv.org/abs/2310.09291},
    doi = {10.48550/arXiv.2310.09291},
    urldate = {2026-01-29},
    publisher = {arXiv},
    author = {Karthik, Shyamgopal and Roth, Karsten and Mancini, Massimiliano and Akata, Zeynep},
    month = feb,
    year = {2024},
    note = {arXiv:2310.09291 [cs]},
}

@article{wang_generative_nodate,
    title = {Generative {Zero}-{Shot} {Composed} {Image} {Retrieval}},
    language = {en},
    author = {Wang, Lan and Ao, Wei and Boddeti, Vishnu Naresh and Lim, Ser-Nam},
}

@misc{ceylan_pix2video_2023,
    title = {{Pix2Video}: {Video} {Editing} using {Image} {Diffusion}},
    shorttitle = {{Pix2Video}},
    url = {http://arxiv.org/abs/2303.12688},
    doi = {10.48550/arXiv.2303.12688},
    urldate = {2026-01-29},
    publisher = {arXiv},
    author = {Ceylan, Duygu and Huang, Chun-Hao Paul and Mitra, Niloy J.},
    month = mar,
    year = {2023},
    note = {arXiv:2303.12688 [cs]},
}

@article{tong_eyes_2024,
    title = {Eyes {Wide} {Shut}? {Exploring} the {Visual} {Shortcomings} of {Multimodal} {LLMs}},
    language = {en},
    author = {Tong, Shengbang and Liu, Zhuang and Zhai, Yuexiang and Ma, Yi and LeCun, Yann and Xie, Saining},
    year = {2024},
}

@inproceedings{karamcheti2024prismatic,
  title={Prismatic vlms: Investigating the design space of visually-conditioned language models},
  author={Karamcheti, Siddharth and Nair, Suraj and Balakrishna, Ashwin and Liang, Percy and Kollar, Thomas and Sadigh, Dorsa},
  booktitle={Forty-first International Conference on Machine Learning},
  year={2024}
}

\newpage
\appendix
\crefalias{section}{appendix}

\section{Models}\label{app:models-section}
For reproducibility, all models are listed in \cref{app:models}. For OpenCLIP models we report the model name and pretrained tag as passed to \texttt{open\_clip.create\_model\_and\_transforms}; for timm models we report the model name as passed to \texttt{timm.create\_model}.
\begin{longtable}{lll}
    \caption{Full list of models used in our study.}
    \label{app:models}\\
    \toprule
    \textbf{Model} & \textbf{Source} & \textbf{Tag} \\
    \midrule
    \multicolumn{3}{l}{\textit{CLIP}} \\
    \midrule
    RN50 (OpenAI)        & OpenCLIP & \texttt{(RN50, openai)} \\
    RN101 (OpenAI)       & OpenCLIP & \texttt{(RN101, openai)} \\
    RN50x4 (OpenAI)      & OpenCLIP & \texttt{(RN50x4, openai)} \\
    ViT-B/32 (LAION-400M)& OpenCLIP & \texttt{(ViT-B-32, laion400m\_e31)} \\
    ViT-B/16 (OpenAI)    & OpenCLIP & \texttt{(ViT-B-16, openai)} \\
    ViT-B/16 (LAION-2B)  & OpenCLIP & \texttt{(ViT-B-16, laion2b\_s34b\_b88k)} \\
    ViT-L/14 (OpenAI)    & OpenCLIP & \texttt{(ViT-L-14, openai)} \\
    ViT-L/14 (LAION-2B)  & OpenCLIP & \texttt{(ViT-L-14, laion2b\_s32b\_b82k)} \\
    ViT-L/14 (CommonPool)& OpenCLIP & \texttt{(ViT-L-14, commonpool\_xl\_s13b\_b90k)} \\
    ViT-L/14 (DataComp)  & OpenCLIP & \texttt{(ViT-L-14, datacomp\_xl\_s13b\_b90k)} \\
    ViT-H/14 (LAION-2B)  & OpenCLIP & \texttt{(ViT-H-14, laion2b\_s32b\_b79k)} \\
    ViT-bigG/14 (LAION-2B)& OpenCLIP & \texttt{(ViT-bigG-14, laion2b\_s39b\_b160k)} \\
    \midrule
    \multicolumn{3}{l}{\textit{SigLIP}} \\
    \midrule
    ViT-B/16             & OpenCLIP & \texttt{(ViT-B-16-SigLIP, webli)} \\
    ViT-L/16-256         & OpenCLIP & \texttt{(ViT-L-16-SigLIP-256, webli)} \\
    ViT-B/16 SigLIP2     & OpenCLIP & \texttt{(ViT-B-16-SigLIP2, webli)} \\
    \midrule
    \multicolumn{3}{l}{\textit{DINO / DINOv2 / DINOv3}} \\
    \midrule
    ViT-S/16 (DINO)      & timm & \texttt{vit\_small\_patch16\_224.dino} \\
    ViT-B/16 (DINO)      & timm & \texttt{vit\_base\_patch16\_224.dino} \\
    ViT-B/8 (DINO)       & timm & \texttt{vit\_base\_patch8\_224.dino} \\
    ViT-B/14 (DINOv2)    & timm & \texttt{vit\_base\_patch14\_dinov2.lvd142m} \\
    ViT-L/14 (DINOv2)    & timm & \texttt{vit\_large\_patch14\_dinov2.lvd142m} \\
    ViT-g/14 (DINOv2)    & timm & \texttt{vit\_giant\_patch14\_dinov2.lvd142m} \\
    ViT-S/16 (DINOv3)    & timm & \texttt{vit\_small\_patch16\_dinov3.lvd1689m} \\
    ViT-B/16 (DINOv3)    & timm & \texttt{vit\_base\_patch16\_dinov3.lvd1689m} \\
    ViT-L/16 (DINOv3)    & timm & \texttt{vit\_large\_patch16\_dinov3.lvd1689m} \\
    \midrule
    \multicolumn{3}{l}{\textit{Masked Image Modeling}} \\
    \midrule
    ViT-B/16 (MAE)       & timm & \texttt{vit\_base\_patch16\_224.mae} \\
    ViT-L/16 (MAE)       & timm & \texttt{vit\_large\_patch16\_224.mae} \\
    ViT-H/14 (MAE)       & timm & \texttt{vit\_huge\_patch14\_224.mae} \\
    BEiT-B/16            & timm & \texttt{beit\_base\_patch16\_224.in22k\_ft\_in22k\_in1k} \\
    BEiT-L/16            & timm & \texttt{beit\_large\_patch16\_224.in22k\_ft\_in22k\_in1k} \\
    BEiTv2-B/16          & timm & \texttt{beitv2\_base\_patch16\_224.in1k\_ft\_in22k\_in1k} \\
    BEiTv2-L/16          & timm & \texttt{beitv2\_large\_patch16\_224.in1k\_ft\_in22k\_in1k} \\
    \midrule
    \multicolumn{3}{l}{\textit{Supervised}} \\
    \midrule
    ResNet-18            & timm & \texttt{resnet18.a1\_in1k} \\
    ResNet-50            & timm & \texttt{resnet50.a1\_in1k} \\
    ResNet-101           & timm & \texttt{resnet101.a1\_in1k} \\
    EfficientNet-B0      & timm & \texttt{efficientnet\_b0.ra\_in1k} \\
    EfficientNet-B4      & timm & \texttt{efficientnet\_b4.ra2\_in1k} \\
    ConvNeXt-B           & timm & \texttt{convnext\_base.fb\_in22k\_ft\_in1k} \\
    ConvNeXt-L           & timm & \texttt{convnext\_large.fb\_in1k} \\
    ConvNeXt-XL          & timm & \texttt{convnext\_xlarge.fb\_in22k\_ft\_in1k} \\
    ViT-S/16             & timm & \texttt{vit\_small\_patch16\_224.augreg\_in21k\_ft\_in1k} \\
    ViT-B/16             & timm & \texttt{vit\_base\_patch16\_224.augreg2\_in21k\_ft\_in1k} \\
    ViT-L/16             & timm & \texttt{vit\_large\_patch16\_224} \\
    TinyViT-21M          & timm & \texttt{tiny\_vit\_21m\_224.dist\_in22k\_ft\_in1k} \\
    DeiT-B/16            & timm & \texttt{deit\_base\_patch16\_224.fb\_in1k} \\
    Swin-T               & timm & \texttt{swin\_tiny\_patch4\_window7\_224.ms\_in1k} \\
    Swin-B               & timm & \texttt{swin\_base\_patch4\_window7\_224.ms\_in22k\_ft\_in1k} \\
    \bottomrule
\end{longtable}

\end{document}